\documentclass[conference]{IEEEtran}
\usepackage[shortlabels]{enumitem}
\usepackage[nospace,noadjust]{cite}
\usepackage{float}
\usepackage{graphicx} 
\usepackage{siunitx}
\usepackage{epsfig} 
\usepackage{amsmath,amsthm,amssymb}  
\usepackage{float}
\usepackage{cases,setspace,adjustbox}
\usepackage{array}
\usepackage[update,prepend]{epstopdf}
\usepackage{caption}
\usepackage{eqparbox}
\usepackage{url}
\usepackage{caption, subcaption}
\usepackage{xcolor}
\usepackage{mathtools}
\usepackage{bbm}
\usepackage{multirow}
\usepackage{multicol}

\usepackage[acronym]{glossaries}
\usepackage{comment}
\usepackage{xcolor}
\newcommand*\circled[1]{\kern-2.5em%
  \put(0,4){\color{white}\circle*{11}}\put(0,4){\circle{11}}%
  \put(-3,0){\color{black}\large#1}~~}
\setacronymstyle{long-short}

\newacronym{5G}{5G}{fifth-generation}
\newacronym{6G}{6G}{sixth-generation}

\newacronym{ae}{AE}{autoencoder}

\newacronym{bler}{BLER}{Block Error Rate}
\newacronym{ci}{CI}{confidence interval}

\newacronym{cu}{CU}{Centralized Unit}
\newacronym{semcs}{SEM-CS}{Statistical Ensemble Method Cold-Start}

\newacronym{dl}{DL}{deep learning}
\newacronym{dqn}{DQN}{Deep Q-Network}
\newacronym{du}{DU}{Distributed Unit}

\newacronym{fa}{FA}{false alarm}

\newacronym{gt}{GT}{ground-truth}
\newacronym{iforest}{IF}{Isolation Forest}

\newacronym{kpi}{KPI}{Key Performance Indicator}

\newacronym{loo}{LOO}{leave-one-out}
\newacronym{lstm}{LSTM}{Long Short-Term Memory}
\newacronym{lstmae}{LSTM-AE}{Long Short-Term Memory autoencoder}
\newacronym{lupi}{LUPI}{learning using privileged information}

\newacronym{mad}{MAD}{Median Absolute Deviation}
\newacronym{mcs}{MCS}{Modulation and Coding Scheme}
\newacronym{mdp}{MDP}{Markov Decision Process}
\newacronym{ml}{ML}{machine learning}

\newacronym{nearrt}{Near-RT}{Near-Real-Time}
\newacronym{nonrt}{Non-RT}{Non-Real-Time}

\newacronym{oran}{O-RAN}{Open Radio Access Network}

\newacronym{ppo}{PPO}{Proximal Policy Optimization}
\newacronym{prb}{PRB}{Physical Resource Block}

\newacronym{ran}{RAN}{Radio Access Network}
\newacronym{rapp}{rApp}{Radio Access Network (RAN) Application}
\newacronym{rf}{RF}{Radio Frequency}
\newacronym{ric}{RIC}{RAN Intelligent Controller}
\newacronym{rl}{RL}{reinforcement learning}
\newacronym{rsrp}{RSRP}{Reference Signal Received Power}
\newacronym{rssi}{RSSI}{Received Signal Strength Indicator}
\newacronym{ru}{RU}{Radio Unit}

\newacronym{sb3}{SB3}{Stable-Baselines3}
\newacronym{sem}{SEM}{Statistical Ensemble Method}
\newacronym{smo}{SMO}{Service Management and Orchestration}
\newacronym{snr}{SNR}{Signal-to-Noise Ratio}
\newacronym{ssb}{SSB}{Synchronization Signal Block}

\newacronym{ue}{UE}{User Equipment}

\newacronym{xapp}{xApp}{eXtended Application}

\begin{document}
\title{CORA: Teacher-Guided Causal O-RAN Adaptation for Jamming Detection}
\author{Sneihil Gopal$^{\ddagger\dagger}$ and Richard A. Rouil$^{\star}$\\
$^{\ddagger}$PREP Associate, National Institute of Standards and Technology (NIST), USA\\
$^{\dagger}$Department of Physics, Georgetown University, USA\\
$^{\star}$National Institute of Standards and Technology (NIST), USA.\\
Emails: \{sneihil.gopal, richard.rouil\}@nist.gov}
\maketitle
\begin{abstract}
In mission-critical networks, a missed jamming event is costlier than a false alarm, so detectors should favor recall, operate causally, and work without labeled attack data. Retrospective Key Performance Indicator (KPI) detectors meet the label-free requirement but rely on future context and cannot run in streaming operation. We present CORA, a teacher-guided reinforcement learning (RL) framework that transfers the hindsight of a retrospective detector to a causal one without labels during training or deployment. Offline, a retrospective KPI detector evaluated on unlabeled historical logs defines a recall-weighted reward for a masked controller that adapts the sensitivity, persistence, and grace of a causal KPI detector. At deployment, the teacher is removed and the frozen controller acts causally, with no ground-truth labels and no target-specific calibration. The design maps onto offline training in the Non-Real-Time RIC and causal control in the Near-Real-Time RIC of an Open Radio Access Network (O-RAN). Under leave-one-out transfer across ten 5G O-RAN testbed configurations
, CORA reaches recall-oriented operating points on the deployable frontier that none of the evaluated baselines covers. A Proximal Policy Optimization (PPO) controller raises mean recall from $\mathbf{0.687}$ to $\mathbf{0.771}$ over a fixed default detector on all ten configurations, while a Deep Q-Network (DQN) achieves a comparable F2-score to the best teacher-selected static configuration, with $\mathbf{61\,\%}$ fewer false alarms. Removing the teacher lowers recall on every configuration.
\end{abstract}
\section{Introduction}
\label{sec:intro}
\gls{rf} jamming threatens network reliability by disrupting legitimate transmissions and degrading communication quality. Detecting it is difficult because its observable \gls{kpi} signatures vary with jammer gain, burst duration, traffic, and the number of active \glspl{ue}. In an \gls{oran} deployment, telemetry exposed through the \gls{ric} enables jamming detection directly from network measurements rather than dedicated spectrum instrumentation~\cite{polese2023understanding,kryszkiewicz2023open,hoffmann2026ocudu}. In mission-critical settings such as command and control, public safety, and surveillance, an undetected jamming event is typically far costlier than a false alarm, so detectors for these settings should favor recall.

A key obstacle is that representative labeled attack data are rarely available for a new deployment. Supervised \gls{ml} detectors require such labels, and statistical detectors depend on fixed operating rules or calibration with target telemetry~\cite{jere2023bayesian}. Our earlier \gls{semcs} detector avoids labels altogether by combining physics-based statistics of \gls{kpi} telemetry~\cite{gopal2026oran}, but it relies on retrospective processing with centered windows, that is, on future context that a streaming detector does not have. To quantify the cost of causality, we constructed a causal variant of \gls{semcs} by replacing these operations with trailing statistics, and its mean F2-score, which prioritizes recall over precision~\cite{vanrijsbergen1979information}, drops from $0.892$ to $0.677$ on our testbed corpus. This raises the question addressed here. Can the hindsight of a retrospective \gls{kpi} detector, applied to historical unlabeled logs, teach a causal detector how to adapt when it is transferred to a held-out configuration?

We answer this question with CORA, a teacher-guided \gls{rl} framework for causal detector adaptation. Offline, \gls{sem}-CS acts as a privileged teacher on historical training logs, in the spirit of \gls{lupi}~\cite{vapnik2009lupi}, and its decisions define a recall-weighted reward. A masked controller learns to adjust the sensitivity, persistence, and grace of a causal \gls{kpi} detector and to release alarms. At deployment the policy is frozen, the teacher is removed, and the controller acts only on current and past observations. 
Although \gls{gt} labels are available for the testbed logs, they are used only for evaluation and never in training or inference. The design maps naturally onto offline training in the \gls{nonrt} \gls{ric} and causal control in the \gls{nearrt} \gls{ric}.

The main contributions are threefold. First, we introduce a teacher-student formulation in which a noncausal detector, evaluated only on unlabeled historical logs, supervises the reward of a causal detector-adaptation policy without \gls{gt} labels or target-specific calibration. Second, we design an incremental masked controller whose mask encodes only stability constraints, train it from random starting configurations, and instantiate it with two learners, \gls{ppo} and \gls{dqn}. Third, we evaluate CORA under \gls{loo} transfer over ten heterogeneous \gls{5G} \gls{oran} testbed configurations 
with five seeds, a pre-specified recall-weighted F2-score, paired statistics, ablations, and an oracle static frontier. Against two static baselines that keep a single detector configuration throughout, Fixed-Default (the detector's default configuration) and Hist-Best-Fixed (the configuration with the highest teacher reward on the training logs), CORA reaches recall-oriented operating points on the deployable frontier that neither baseline covers. \gls{ppo} raises mean recall from $0.687$ to $0.771$ over Fixed-Default on every held-out configuration, and \gls{dqn} matches Hist-Best-Fixed in F2-score with $61\,\%$ fewer false alarms. 
The \gls{sem}-CS teacher and an oracle static frontier scored with GT labels serve as hindsight references rather than deployable alternatives. CORA's contribution is to approach their recall-oriented operating points using only unlabeled historical logs. The remainder of this paper is organized as follows. Section~\ref{sec:related_work} reviews related work, Section~\ref{sec:method} presents CORA, Section~\ref{sec:methodology} describes the experimental methodology, Section~\ref{sec:results} presents the results, and Section~\ref{sec:conclusion} concludes the paper.
\section{Related Work}
\label{sec:related_work}
\gls{rf} jamming detection has been studied with statistical, anomaly-detection, and learning-based techniques, using features from physical-layer measurements to network-level telemetry~\cite{pirayesh2022jamming,jere2023bayesian}. \gls{oran} exposes such telemetry through the \gls{ric}~\cite{polese2023understanding}, and prior work has demonstrated jamming detection in \gls{5G} \gls{oran} testbeds and recent work on deployable implementations~\cite{kryszkiewicz2023open,hoffmann2026ocudu,gopal2026oran}, with detection and adaptive response identified as key functions for resilient operation~\cite{gopal2025adversary_resilient}. Our earlier \gls{semcs} detector combines physics-based \gls{kpi} statistics without labeled training data~\cite{gopal2026oran}, but its centered windows and retrospective post-processing prevent causal operation. Unsupervised anomaly detectors apply a fixed decision rule after training, and supervised detectors require labeled attacks that are rarely available for a new deployment. None of these approaches adapts the operating point of a causal detector online, under a label-free deployment constraint, when it is transferred to a new configuration.

\gls{rl} is widely used for closed-loop \gls{oran} control, where \glspl{xapp} and \gls{rapp} adjust \acrshort{ran} parameters from streaming \glspl{kpi}~\cite{bonati2021intelligence}. These controllers typically optimize directly observable rewards such as throughput or latency, but jamming presence is unobservable at the controller and cannot serve directly as a reward without labels. \gls{lupi}~\cite{vapnik2009lupi} and knowledge distillation~\cite{hinton2015distilling} address related settings using information or a teacher model available only during training. CORA departs from both: the privileged information is future context from a retrospective noncausal detector evaluated on unlabeled historical logs, and the student is an \gls{rl} controller that adapts a causal detector rather than a classifier imitating the teacher. Because the teacher enters only through the reward, the student can weight recall above agreement and requires neither labels nor the teacher at deployment.
\section{Teacher-Guided Causal Detector Adaptation}
\label{sec:method}

CORA learns to adapt the operating point of a causal \gls{kpi}-based jamming detector. Training uses historical telemetry only, with a retrospective statistical detector operating on the same KPI telemetry as an offline teacher. 
Here the privileged information is future context. At deployment the teacher is removed, and the controller acts on current and past observations only. \gls{gt} labels are not used in teacher generation, reward computation, policy training, or causal inference, and are reserved solely for evaluation. Next, we describe the three components of CORA: the causal detector, the teacher-guided control, and training and deployment.

\textbf{Causal detector:}
Let $x_{i,t}$ denote \gls{kpi} $i\in\{1,\dots,N\}$ at sample $t$, where the $N=9$ \glspl{kpi} are six \gls{ssb}-\gls{rsrp} statistics, namely the mean, minimum, three quartiles, and maximum, together with the Count of \gls{ssb}-\gls{rsrp} measurements, \gls{snr}, and \gls{rssi}. The robust level deviation of \gls{kpi} $i$ is measured against the median and \gls{mad} of a trailing window of $W=30$ samples ending at $t$,
\begin{equation}
z_{i,t}=\frac{|x_{i,t}-\operatorname{med}_W(x_i)|}{\operatorname{MAD}_W(x_i)},
\label{eq:robust_z}
\end{equation}
and the change deviation $z^{\Delta}_{i,t}$ is defined in the same way on the absolute first difference $|x_{i,t}-x_{i,t-1}|$. The Count deviation $z^{\mathrm{C}}_t$ measures how far Count, the number of completed measurements in a report, falls below its trailing median. The sensitivity level $k\in\{0,\dots,6\}$ selects three thresholds $(\zeta_k,\zeta^{\Delta}_k,v_k)$, where $\zeta_k$ is the threshold on the level deviations, $\zeta^{\Delta}_k$ is the threshold on the change deviations, and $v_k$ is the threshold on the detector confidence defined next. All three decrease monotonically in $k$, so that a larger $k$ yields a more sensitive detector. With $\mathbf{1}[\cdot]$ the indicator function, the detector confidence is
\begin{equation}
\begin{split}
h_t={}&\frac{0.40}{N}\sum_{i=1}^{N}\mathbf{1}[z_{i,t}>\zeta_k]+\frac{0.25}{N}\sum_{i=1}^{N}\mathbf{1}[z^{\Delta}_{i,t}>\zeta^{\Delta}_k]\\
&\hspace{13em}+0.35\,\phi(z^{\mathrm{C}}_t),
\end{split}
\label{eq:confidence}
\end{equation}
that is, a weighted combination of the fractions of \glspl{kpi} with level and change deviations above threshold and a Count-drop term, with weights set as in~\cite{gopal2026oran} and $\phi(\cdot)$ mapping a deviation to $[0,1]$. A raw detection occurs when $h_t\geq v_k$ or Count equals zero. An alarm is raised after $p\in\{2,\dots,6\}$ consecutive raw detections, or immediately when Count equals zero. To bridge short gaps within an alarm, the detector uses a causal evidence signal $e_t=\max(u_t,d_t)\in[0,1]$, where the unavailability indicator $u_t$ equals one when Count is zero or the \gls{rsrp} report is missing and zero otherwise, and the directional evidence $d_t\in[0,1]$ measures how far \gls{snr} has decreased and \gls{rssi} has increased relative to a lagged reference window, as expected under jamming. While an alarm is active, the grace parameter $g\in\{0,\dots,4\}$ lets it bridge up to $g$ raw-negative samples when $e_t$ remains high. The detector configuration is thus $(k,p,g)$, with default $(3,2,0)$.

\textbf{Teacher-guided control:}
The teacher is the \gls{semcs} detector~\cite{gopal2026oran}, whose parameters are derived from signal physics rather than labels. It uses centered windows and retrospective post-processing and therefore cannot run causally. It is evaluated per sample on each historical training log to produce a binary decision $q_t\in\{0,1\}$, and it is never evaluated on the held-out configuration.

Every block of $B=5$ samples, the controller selects one incremental action from
\begin{equation}
\mathcal{A}=\{\mathrm{hold},\,k^{\pm},\,p^{\pm},\,g^{\pm},\,\mathrm{release}\},
\label{eq:actions}
\end{equation}
where $k^{\pm}$, $p^{\pm}$, and $g^{\pm}$ raise or lower sensitivity, persistence, or grace by one step, and release ends an active alarm without changing $(k,p,g)$. The block interval amortizes control decisions over several \gls{kpi} reports, giving the controller time to observe the effect of a configuration change before taking another. An action mask removes moves beyond the parameter bounds, enforces a minimum dwell of two blocks between configuration changes, and allows release only during an active alarm when Count is non-zero. The mask encodes stability constraints only, so the direction of adaptation is learned rather than prescribed. Let $\mathbf{s}_b$ denote the $34$-dimensional controller state at control block $b$. It contains the current, short-term ($5$ samples), and long-term ($20$ samples) values of the evidence $e_t$, its trend, variability, and unavailability rate, the normalized $(k,p,g)$, recent alarm and raw-detection rates, the alarm status and age, the blocks since the last change, the Count deviation, the previous action, and the level and change deviations $z_{i,t}$ and $z^{\Delta}_{i,t}$ of all $N$ \glspl{kpi}.

The reward compares the causal alarm state $\hat{y}_t\in\{0,1\}$ with the teacher decision $q_t$ as
\begin{equation}
\begin{split}
r_t={}&\hat{y}_t\left(q_t-(1-q_t)\right)\\
&+(1-\hat{y}_t)\left(0.5\,(1-q_t)-1.5\,q_t\right)-0.25\,o_t,
\end{split}
\label{eq:reward}
\end{equation}
where $o_t$ indicates an alarm onset. Missing a teacher-positive sample therefore costs $1.5$ times a teacher-negative alarm, reflecting the higher operational cost of undetected jamming, while the onset term discourages fragmented alarms. The $1.5:1$ miss-to-false-alarm penalty reflects the same recall-oriented objective as the F2-score and was fixed a priori. The onset and change penalties were set to the smallest values that visibly suppressed alarm fragmentation and oscillation during early training. The block reward is the mean of $r_t$ over the block, minus $0.02$ per configuration change.

\begin{figure}[t]
    \centering    
    \includegraphics[width=0.9\columnwidth]
    {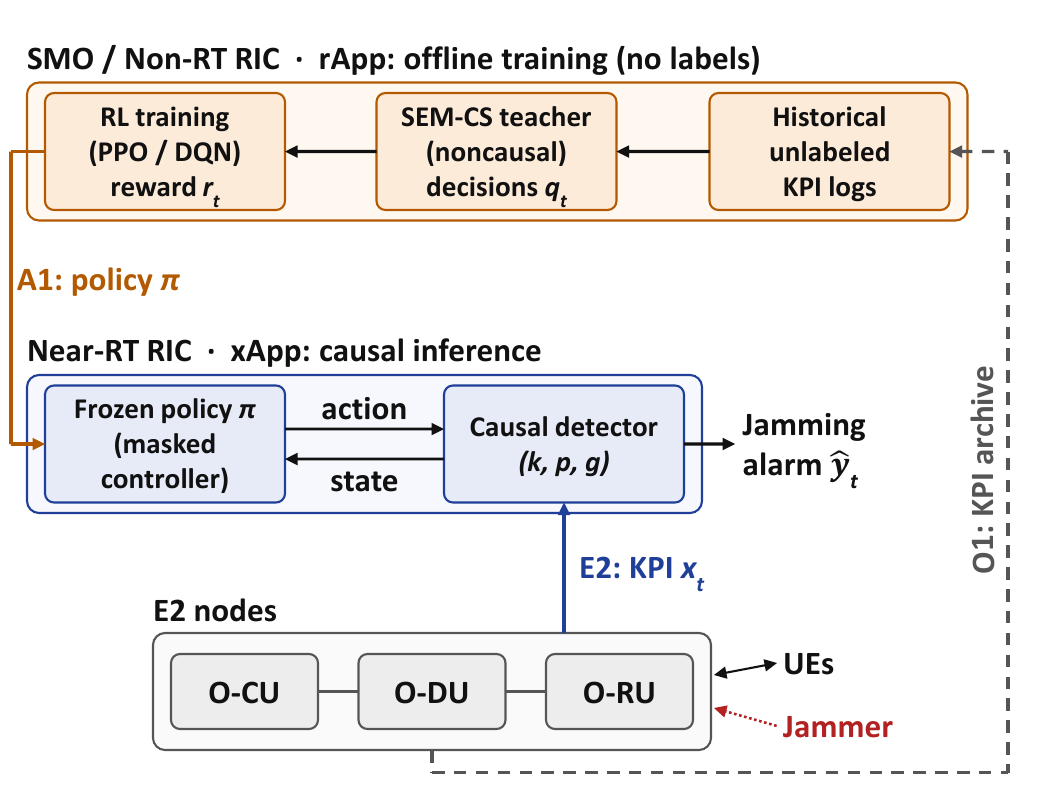}
    \caption{\acrshort{oran} deployment of CORA.}
    \label{fig:oran_deployment}
\end{figure}
\textbf{Training and deployment:}
Each training episode spans $128$ blocks of a uniformly sampled historical log and begins at a uniformly sampled position after a $120$-sample warm-up. The episode starts from a uniformly random $(k,p,g)$, so that the policy learns how to move from any configuration rather than only from the default. The primary learner is \gls{ppo}~\cite{schulman2017ppo} with invalid-action masking~\cite{huang2022masking}, using two $64$-unit layers, learning rate $3\times10^{-4}$, rollout length $1024$, batch size $128$, $\gamma=0.99$, $\lambda=0.95$, clip range $0.2$, entropy coefficient $0.01$, and $10$ epochs per update. To test whether the results depend on the learning algorithm, we also train \gls{dqn}~\cite{mnih2015dqn} as a second learner under the same detector state, action space, mask constraints, and reward. \gls{dqn} receives the action mask as part of its input and excludes masked actions from greedy selection, bootstrap targets, and $\epsilon$-greedy exploration. It uses learning rate $10^{-4}$, a replay buffer of $50{,}000$ transitions, batch size $32$, updates every $4$ steps, target updates every $1{,}000$ steps, and $\epsilon$ decayed from $1.0$ to $0.05$ over the first $10\%$ of training. These settings follow those of the original algorithms~\cite{schulman2017ppo,mnih2015dqn}, with the rollout length, replay buffer, and target-update interval reduced for the short training budget, and none were tuned on held-out data. Both learners are trained for $\num{50000}$ control steps, a budget selected using the teacher-reward training return only. At deployment, the frozen policy starts from the default configuration and acts causally.

As shown in Figure~\ref{fig:oran_deployment}, CORA separates naturally across the two \gls{ric} control loops. In the \gls{nonrt} \gls{ric}, an \gls{rapp} evaluates the retrospective teacher on archived \gls{kpi} logs, trains the controller, and delivers the frozen policy to the \gls{nearrt} \gls{ric} over the A1 interface. In the \gls{nearrt} \gls{ric}, an \gls{xapp} receives \gls{kpi} reports over the E2 interface, updates the causal detector at each report, and applies a controller action every $B=5$ reports. The deployed policy is a two-layer network with $64$ units per layer, so inference adds negligible computation, and neither the teacher nor any label is required online. The same artifact ships whether the deployment has labeled attack data or not. As new unlabeled logs accumulate, the \gls{nonrt} \gls{ric} can periodically retrain the policy without interrupting causal detection.
\section{Experimental Methodology}
\label{sec:methodology}

This section describes the testbed corpus and transfer protocol, the reference methods, and the metrics and statistical tests used to evaluate CORA.

\textbf{Corpus and protocol:}
The corpus consists of ten logs recorded on a \gls{5G} \gls{oran} testbed, one per testbed configuration, with a total duration of $34.7$\,h (Table~\ref{tab:corpus}). Logs~$1$ to~$8$ use a constant jammer gain with fixed or variable burst durations, whereas Logs~$9$ and~$10$ introduce short irregular bursts with dynamic gain. Each sample contains the nine \glspl{kpi} defined in Section~\ref{sec:method}, among which Count reflects measurement availability during jamming. Each \gls{ue} reports once per second, and the reports of all \glspl{ue} in a configuration are processed in recorded order as a single causal stream, modeling the cell as the unit of detection; per-\gls{ue} processing is left to future work.

\begin{table}[t]
    {\small
     \vspace*{0.08in}
    \centering
    \begin{tabular}{p{0.35cm}p{0.25cm}p{1cm}p{1cm}p{0.75cm}p{1.35cm}p{1cm}}
    \hline
    \textbf{Log} & \textbf{\acrshort{ue}s} & \textbf{Duration} &
    \textbf{Burst (s)} & \textbf{Gain (dB)} & \textbf{Burst Pattern} &
    \textbf{Samples} \\
    \hline \hline
    $1$ & $3$ & $68$\,min   & $60$       & $48$ & Fixed    & \num{12186} \\
    $2$ & $2$ & $62$\,min   & $60$       & $48$ & Fixed    & \num{7468}  \\
    $3$ & $2$ & $43$\,min & $60$       & $48$ & Fixed    & \num{5132}  \\
    $4$ & $1$ & $47$\,min   & $60$       & $48$ & Fixed    & \num{2811}  \\
    $5$ & $1$ & $79$\,min   & $30$       & $48$ & Fixed    & \num{4750}  \\
    $6$ & $1$ & $15$\,min & $60$       & $45$ & Fixed    & \num{926}      \\
    $7$ & $1$ & $3.1$\,h   & $[10,30]$  & $45$ & Variable & \num{10981} \\
    $8$ & $1$ & $18.1$\,h  & $[10,30]$  & $45$ & Variable & \num{65075} \\
    $9$ & $3$ & $2.0$\,h  & $[5,14]$  & $[48,60]$ & Variable & \num{17465} \\
    $10$ & $3$ & $6.3$\,h  & $[5,14]$  & $[48,60]$ & Variable & \num{59028} \\
    \hline
    \end{tabular}
    \caption{Testbed configurations.}
    \label{tab:corpus}}
\end{table}

We use ten-fold \gls{loo} transfer. In each fold, one configuration is held out, and the retrospective teacher is evaluated only on the remaining nine, which provide the training experience. The held-out configuration is processed causally from the default detector configuration, with no teacher, retraining, or target-specific calibration. \gls{gt} labels are loaded only after the held-out predictions are generated. \gls{ppo} and \gls{dqn} are each trained for $50{,}000$ control steps with five seeds, and metrics are averaged across seeds within each fold before aggregation; statistical power comes from the per-fold paired comparison across ten configurations rather than seed variance. The ten configurations were used during method development, so although all design choices were frozen before the reported runs, they do not constitute an untouched validation set.

\textbf{Reference methods:}
We compare CORA with five groups of reference methods. Two static baselines keep a single configuration throughout: \textit{Fixed-Default} uses $(k,p,g)=(3,2,0)$, and \textit{Hist-Best-Fixed} uses the configuration among the $175$ combinations of $(k,p,g)$ with the highest teacher reward on the nine training logs. This is deployable but requires that the training logs represent the deployment conditions, and it ships a single globally-selected configuration rather than one adapted per deployment. \textit{Random+Mask} selects uniformly among the same masked actions as the learners, testing whether the policy is actually learned. For existing causal detection we include \textit{Causal-\gls{semcs}}, the causal re-engineering of \gls{semcs} from Section~\ref{sec:intro} that replaces centered windows with trailing statistics, \textit{Adaptive-Hyst}, a hand-designed hysteresis controller over $(k,p,g)$, and \textit{Count-Zero}, which alarms whenever Count equals zero. Three \gls{ppo} ablations each remove one design choice: random training starts (replaced by always starting from the default), the dwell-only mask (replaced by an evidence-gated mask), and the teacher (replaced by a state-evidence reward). Finally, two non-deployable references bound what is achievable: the \gls{semcs} teacher shows what retrospective processing achieves, and an \textit{oracle frontier} scores all $175$ static configurations on each held-out log with \gls{gt} labels.

\textbf{Metrics and statistics:}
The pre-specified primary metric is the \textit{F2-score}, the $\beta=2$ case of the F-measure $(1+\beta^2)PR/(\beta^2P+R)$~\cite{vanrijsbergen1979information}, which weights recall above precision and reflects the higher cost of missed jamming. We also report F1-score, precision, recall, \textit{event recall}, \textit{false-alarm intervals} per $\num{1000}$ samples (\acrshort{fa}/$1$k), and \textit{detection delay}. A jamming event is a contiguous \gls{gt}-positive interval, counted as detected if any alarm interval overlaps it; a false alarm is an alarm interval that overlaps no event; delay is the time from event onset to the first overlapping alarm, averaged over detected events. Metrics are computed per held-out configuration and macro-averaged, with pooled sample-weighted metrics as a secondary view. Comparisons use per-fold paired differences with two-sided Wilcoxon signed-rank tests~\cite{demsar2006statistical} and paired bootstrap $95\,\%$ \glspl{ci} from $\num{10000}$ resamples. Holm correction~\cite{demsar2006statistical} is applied separately for each learner and metric.
\section{Results and Discussion}
\label{sec:results}
Table~\ref{tab:main_results} reports macro-averaged held-out performance across the ten \gls{loo} folds. We first ask whether CORA improves detection and whether its policy is actually learned, then identify where the gain comes from, and finally examine operating points, controller behavior, and the conditions under which adaptation helps.

\begin{table*}[t]
\centering
 \vspace*{0.08in}
{\small
\begin{tabular}{l|cccccccc}
\hline
Method & Causal & F2-score & F1-score & Precision & Recall & Event recall & \acrshort{fa}/$1$k & Delay (s) \\
\hline
\acrshort{sem}-CS teacher (offline reference) & No & $0.892$ & $0.830$ & $0.753$ & $0.945$ & $0.979$ & $21.63$ & $0.30$ \\
\hline
\textbf{\acrshort{ppo} (proposed)} & Yes & $\mathbf{0.742}$ & $0.715$ & $0.735$ & $0.771$ & $0.979$ & $24.93$ & $1.15$ \\
\textbf{\acrshort{dqn} (proposed)} & Yes & $0.732$ & $0.733$ & $0.820$ & $0.738$ & $0.969$ & $13.11$ & $1.75$ \\
Hist-Best-Fixed & Yes & $0.734$ & $0.688$ & $0.681$ & $\mathbf{0.784}$ & $\mathbf{0.985}$ & $33.26$ & $0.69$ \\
Fixed-Default & Yes & $0.711$ & $\mathbf{0.758}$ & $\mathbf{0.963}$ & $0.687$ & $0.973$ & $1.98$ & $1.61$ \\
Random+Mask & Yes & $0.689$ & $0.726$ & $0.884$ & $0.670$ & $0.882$ & $4.31$ & $2.56$ \\
Causal-\acrshort{sem}-CS & Yes & $0.677$ & $0.649$ & $0.631$ & $0.704$ & $0.865$ & $8.05$ & $8.03$ \\
Count-Zero & Yes & $0.646$ & $0.705$ & $0.887$ & $0.616$ & $0.818$ & $0.50$ & $2.97$ \\
Adaptive-Hyst & Yes & $0.642$ & $0.667$ & $0.791$ & $0.634$ & $0.830$ & $16.14$ & $2.62$ \\
\hline
\end{tabular}
\caption{Macro-average held-out performance over ten \acrshort{loo} folds}
\label{tab:main_results}}
\end{table*}

\textbf{Overall performance:}
CORA shifts the causal detector to a higher-recall operating point. Relative to Fixed-Default, \gls{ppo} raises recall from $0.687$ to $0.771$ on every held-out configuration (Holm-adjusted $p=0.016$), at the cost of more false alarms and lower precision, which is why Fixed-Default keeps the best F1-score. In mission-critical settings in which a missed attack is costlier than a false alarm, this trade is intended: \gls{ppo}'s additional false alarms buy $8.4$ percentage points of recall ($0.771$ versus $0.687$) that Fixed-Default misses. The F2-scores reflect this trade without a clear winner: \gls{ppo} attains the highest mean F2 among causal methods ($0.742$), but its margins over Hist-Best-Fixed ($0.734$), \gls{dqn} ($0.732$), and Fixed-Default ($0.711$) are not statistically significant. Both learners match or exceed the event recall of all deployable baselines except Hist-Best-Fixed, and \gls{ppo} alarms $6.9$\,s earlier than Causal-\gls{semcs}.

These gains come from learning rather than from the action space itself. Random+Mask 
falls below \gls{ppo} and \gls{dqn} in recall by $0.102$ and $0.068$ on every configuration (Holm-adjusted $p\leq0.016$) and misses many more jamming events.

\textbf{Where the gain comes from:}
CORA combines two ingredients, the retrospective teacher, which provides the training signal, and the adaptation of $(k,p,g)$ during operation. To separate their contributions, we compare three methods that add them one at a time, namely Fixed-Default, which uses neither, Hist-Best-Fixed, which uses only the teacher to select one static configuration, and the CORA controllers \gls{ppo} and \gls{dqn}, which use both. The teacher alone already moves the detector to a higher-recall region. Hist-Best-Fixed raises recall over Fixed-Default by $0.097$ on every configuration ($p=0.002$), but a single static configuration must be sensitive everywhere and therefore raises the most false alarms of all causal methods. Adding adaptation keeps most of this recall at a lower false-alarm cost. \gls{ppo} keeps recall close to that of Hist-Best-Fixed ($0.771$ versus $0.784$) with $25\,\%$ fewer false alarms, although this reduction is not significant after Holm correction, whereas \gls{dqn} trades a little recall for a $61\,\%$ reduction in false alarms on every configuration (Holm-adjusted $p=0.014$). In short, the teacher tells the detector where to operate, and adaptation lets it operate there more efficiently than any single teacher-selected configuration. Both Hist-Best-Fixed and CORA require that the training logs cover the deployment regime, and neither recovers when this fails (Limitations). But within-regime, CORA offers an operational advantage that Hist-Best-Fixed does not. The deployed policy adjusts sensitivity per deployment rather than shipping a single globally-selected configuration to every target.

Comparing the two learners tests whether these results depend on the learning algorithm. Although \gls{ppo} and \gls{dqn} use very different update rules, their F2-scores are statistically indistinguishable and stable across seeds (per-fold seed standard deviation below $0.01$ on nine of ten folds for both learners), indicating that the qualitative behavior is not specific to a particular \gls{rl} algorithm. The learners differ only in where they settle: \gls{dqn} chooses a more conservative operating point and \gls{ppo} a more recall-oriented one.

\begin{table}[t]
\centering
{\small
\begin{tabular}{lccc}
\hline
Variant & F2-score & Recall & \acrshort{fa}/$1$k \\
\hline
Full \acrshort{ppo} & $0.742$ & $0.771$ & $24.93$ \\
No random training start & $0.749$ & $0.794$ & $31.31^\dagger$ \\
Evidence-gated mask & $0.724$ & $0.708$ & $4.51^\dagger$ \\
No teacher (evidence reward) & $0.682$ & $0.655^\dagger$ & $3.34^\dagger$ \\
\hline
\end{tabular}
\caption{\acrshort{ppo} ablations. The dagger ($\dagger$) marks a difference from full \acrshort{ppo} that is in the same direction on all ten folds with Holm-adjusted $p<0.01$.}
\label{tab:ablation}}
\end{table}

\textbf{Ablations:}
Table~\ref{tab:ablation} confirms that the teacher drives the recall gain. Replacing it with a reward computed from causal evidence alone lowers recall on every configuration (Holm-adjusted $p=0.006$) and brings \gls{ppo} close to Fixed-Default, because such a reward only confirms what the causal detector can already observe. The other two design choices mainly decide where the policy settles. Without random training starts, the policy only learns to move from the default and becomes more aggressive, trading a slightly higher F2-score on this corpus for narrower exposure to starting configurations during training. The evidence-gated mask, which permits only evidence-supported moves, keeps the policy conservative. Neither changes the F2-score significantly.

\begin{figure}[t]
    \centering
    \includegraphics[width=\columnwidth]
    {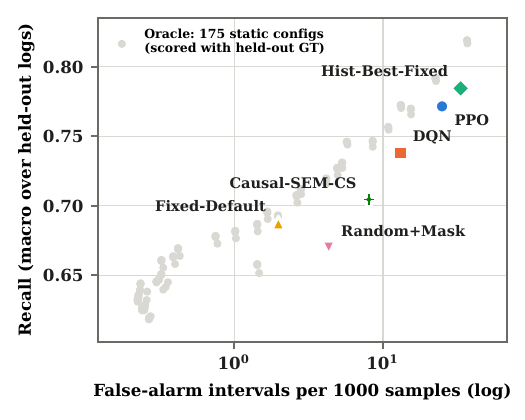}
    \caption{Recall versus \acrshort{fa}/1k. Grey points: oracle frontier of $175$ static configurations scored with held-out \acrshort{gt}.}
    \label{fig:recall_fa}
\end{figure}

\textbf{Operating point and controller behavior:}
Figure~\ref{fig:recall_fa} places all methods in the recall versus false-alarm plane, together with the oracle frontier of static configurations scored with held-out labels. The oracle requires \gls{gt} labels on the target configuration to select the best static $(k,p,g)$ per deployment and is therefore not deployable. It bounds what any static choice could achieve in hindsight. \gls{ppo}, \gls{dqn}, and Hist-Best-Fixed lie on or slightly below this frontier. Among methods that could actually ship, CORA is the only one reaching intermediate false-alarm budgets at near-oracle recall. Hist-Best-Fixed reaches comparable recall only by always running at maximum sensitivity, and Fixed-Default meets strict false-alarm budgets only by missing more jamming. Under a strict budget of $5$ \acrshort{fa}/$1$k, however, Fixed-Default remains the right choice, since it meets this budget on nine configurations, whereas \gls{ppo} meets it on none and \gls{dqn} on one.

\begin{figure}[t]
    \centering
    \includegraphics[width=\columnwidth]
    {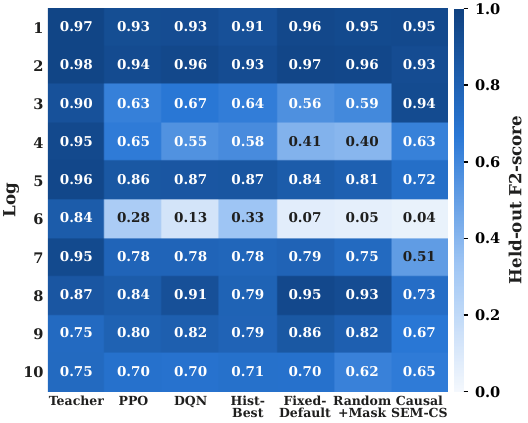}
    \caption{Held-out F2-score per configuration and method (learners: mean of five seeds).}
    \label{fig:per_fold}
\end{figure}

The controllers reach these operating points by adapting to each configuration rather than by applying one fixed setting. Hist-Best-Fixed always uses a high sensitivity, whereas the learners choose a sensitivity level per held-out configuration and then mostly hold it. Measured across all control blocks aggregated over the ten folds, \gls{dqn} changes the configuration in only $8\,\%$ of blocks and Random+Mask in $31\,\%$, while \gls{ppo} lies between the two. \gls{dqn} keeps the sensitivity near the default where Fixed-Default already performs well and raises it where Fixed-Default misses jamming (Figure~\ref{fig:per_fold}), whereas \gls{ppo} stays at high sensitivity on most configurations. Random+Mask, by contrast, changes the configuration without a consistent direction.

\textbf{When does adaptation help?}
Figure~\ref{fig:per_fold} also shows that the benefit of adaptation varies strongly across configurations. A post hoc diagnostic, computed with \gls{gt} after evaluation and not used by the method, explains this through the jamming signature. Where jamming causes the loss of \gls{ssb}-\gls{rsrp} measurements, the default detector already succeeds, and where it does not, detection depends on weaker \gls{snr} and \gls{rssi} shifts that the controllers compensate for by raising sensitivity. Across configurations, the rate of measurement loss tracks both the Fixed-Default F2-score (Spearman $\rho=0.98$) and the recall gain of \gls{ppo} ($\rho=-0.85$); the negative sign reflects that where Fixed-Default already captures jamming events via measurement loss, \gls{ppo} has little room to improve. For each learner and configuration, we define the recall transfer efficiency $\eta=(R_{\mathrm{student}}-R_{\mathrm{fixed}})/(R_{\mathrm{teacher}}-R_{\mathrm{fixed}}).$ Across the seven configurations where $R_{\mathrm{teacher}}-R_{\mathrm{fixed}}>0.05$, \gls{ppo} recovers a median $43\,\%$ of the teacher's advantage and \gls{dqn} $26\,\%$, least on Log~$6$, where jamming never removes measurements and all causal detectors remain weak while the teacher reaches F2 of $0.84$. CORA therefore helps most where jamming is subtle, limited by how much of the teacher's hindsight a causal detector can reproduce.

\textbf{Limitations:}
The averages hide configurations where adaptation does not pay off. Fixed-Default remains strongest on the sample-dominant Logs~$8$ and~$10$, and in pooled sample-weighted metrics (dominated by these logs) its F2-score ($0.836$) exceeds that of both \gls{dqn} ($0.823$) and \gls{ppo} ($0.801$). For deployments resembling these high-sample measurement-loss regimes, Fixed-Default is the better choice, and CORA's advantage is concentrated in the measurement-preserving regimes typified by Logs~$3$ through~$7$. The student also inherits the biases of the teacher. When the learners are trained only on the constant-gain Logs~$1$ to~$8$ and tested on the dynamic-gain Logs~$9$ and~$10$, the teacher becomes less precise, and both the teacher and the learners fall behind Fixed-Default (F2-scores of $0.750$ for the teacher, $0.747$ for \gls{ppo}, and $0.735$ for \gls{dqn}, against $0.782$ for Fixed-Default). Student quality is therefore bounded by teacher quality under distribution shift, so training logs should cover the deployment regimes. Finally, the ten configurations were also used in method development, so although the design was frozen before the reported evaluation, they are not 
an untouched validation set.

\section{Conclusion}
\label{sec:conclusion}
We presented CORA, a teacher-guided \gls{rl} framework in which a retrospective \gls{kpi} detector, evaluated offline on unlabeled historical logs, trains a controller that adapts a causal \gls{kpi} detector without \gls{gt} labels or target-specific calibration. Under \gls{loo} transfer across ten \gls{5G} \gls{oran} testbed configurations, CORA shifted the causal detector to a higher-recall operating point, raising recall over the fixed default on every configuration. The retrospective teacher drives this gain, which adaptation retains at fewer false alarms than any single teacher-selected configuration. \gls{ppo} and \gls{dqn} occupy different operating points on the deployable frontier at indistinguishable F2-scores, with PPO more recall-oriented and DQN more conservative on false alarms. Non-deployable references, the retrospective teacher and the oracle static frontier, bound what is achievable with future context and target-specific labels respectively but are not operational alternatives. Future work includes regime-aware teachers, explicit false-alarm budgets, and per-\gls{ue} processing.
\section*{Acknowledgment}
This work is partially funded by the Department of Homeland Security’s Science and Technology Directorate (S\&T). U.S. Government work, not subject to U.S. Copyright.
\begin{spacing}{}
\bibliographystyle{IEEEtran}
\bibliography{references}
\end{spacing}
\end{document}